\documentclass{article}

    \PassOptionsToPackage{numbers}{natbib}

\usepackage[preprint]{neurips_2024}

\usepackage[utf8]{inputenc} 
\usepackage[T1]{fontenc}    
\usepackage{hyperref}       
\usepackage{url}            
\usepackage{booktabs}       
\usepackage{amsfonts}       
\usepackage{nicefrac}       
\usepackage{microtype}      
\usepackage{xcolor}         
\usepackage{graphicx}
\usepackage{subcaption}
\usepackage{algorithm}
\usepackage{algpseudocode}
\usepackage[most]{tcolorbox}
\usepackage{multirow}
\usepackage{tabularx}

\usepackage{amsmath}
\usepackage[capitalise]{cleveref}

\usepackage{bm}

\newcommand{\xv}{\bm{x}}    
\newcommand{\hv}{\bm{h}}    
\newcommand{\zv}{\bm{z}}    
\newcommand{\ztv}{\tilde{\bm{z}}} 
\newcommand{\sv}{\bm{s}}  
\newcommand{\stv}{\tilde{\bm{s}}} 
\newcommand{\vocab}{\mathcal{V}}    
\newcommand{\Attn}{\mathsf{Attn}}   
\newcommand{\FFN}{\mathsf{FFN}} 
\newcommand{\thetab}{\bm{\theta}}   
\newcommand{\mv}{\bm{m}} 
\newcommand{\E}{\mathbb{E}} 
\newcommand{\ones}{\mathbf{1}}  
\newcommand{\argwhere}{\mathsf{argwhere}}    
\newcommand{\KL}{\mathbb{D}_{KL}}
\newcommand{\softmax}{\mathsf{softmax}}
\newcommand{\uv}{\bm{u}}    
\newcommand{\vv}{\bm{v}}    

\newcommand{\proj}[0]{\textsc{FinerCut}}

\definecolor{attncolor}{RGB}{255, 173, 173}
\definecolor{mlpcolor}{RGB}{160, 196, 255}
\definecolor{trcolor}{RGB}{189, 178, 255}
\definecolor{keepcolor}{RGB}{13, 48, 130}
\definecolor{dropattncolor}{RGB}{255, 194, 71}
\definecolor{dropffncolor}{RGB}{115, 169, 66}

\newtcbox{\attnbox}{on line,
    colframe=attncolor, colback=attncolor!100!white,
    boxrule=0pt, boxsep=0pt, left=1pt, right=1pt, top=2pt, bottom=2pt,
    arc=2pt, outer arc=2pt}

\newtcbox{\mlpbox}{on line,
    colframe=mlpcolor, colback=mlpcolor!100!white,
    boxrule=0pt, boxsep=0pt, left=1pt, right=1pt, top=2pt, bottom=2pt,
    arc=2pt, outer arc=2pt}

\newtcbox{\trbox}{on line,
    colframe=trcolor, colback=trcolor!100!white,
    boxrule=0pt, boxsep=0pt, left=1pt, right=1pt, top=2pt, bottom=2pt,
    arc=2pt, outer arc=2pt}

\title{\proj{}: Finer-grained Interpretable Layer Pruning for Large Language Models }

\author{
        $\text{Yang Zhang }^{\ast 1}$ \: $\text{Yawei Li }^{\ast  2,3}$ \:  $\text{Xinpeng Wang}^{2,3}$ \: $\text{Qianli Shen}^{1}$  \: 
        \\$\textbf{Barbara Plank}^{2,3}$ \:  $\textbf{Bernd Bischl}^{2,3}$ \: $\textbf{Mina Rezaei}^{2,3}$ \: $\textbf{Kenji Kawaguchi}^{1}$\\
        ${}^{1}\text{National University of Singapore}$  \:
        ${}^{2}\text{LMU Munich}$ \\
        ${}^{3}\text{Munich Center for Machine Learning (MCML)}$\\
        }

\begin{document}

\maketitle

\begin{abstract}
Overparametrized transformer networks are the state-of-the-art architecture for Large Language Models (LLMs). However, such models contain billions of parameters making large compute a necessity, while raising environmental concerns.
To address these issues, we propose \proj, a new form of fine-grained layer pruning, which in contrast to prior work at the transformer block level, considers all self-attention and feed-forward network (FFN) layers within blocks as individual pruning candidates. \proj{} prunes layers whose removal causes minimal alternation to the model's output---contributing to a new, lean, interpretable, and task-agnostic pruning method.
Tested across 9 benchmarks, our approach retains $90\%$ performance of Llama3-8B with $25\%$ layers removed, and $95\%$ performance of Llama3-70B with $30\%$ layers removed, all without fine-tuning or post-pruning reconstruction. 
Strikingly, we observe intriguing results with \proj: 42\% ($34$ out of $80$) of the self-attention layers in Llama3-70B can be removed while preserving $99\%$ of its performance---without additional fine-tuning after removal. Moreover, \proj{} provides a tool to inspect the types and locations of pruned layers, allowing to observe interesting pruning behaviors. For instance, we observe a preference for pruning self-attention layers, often at deeper consecutive decoder layers.  We hope our insights inspire future efficient LLM architecture designs.
\let\thefootnote\relax\footnotetext{$^{\ast}$ Equal contribution}
\end{abstract}
\section{Introduction}
Large language models (LLMs) have shown impressive performance improvement in recent years and have exhibited emergent abilities~\citep{radford2019language, brown2020language, wei2021finetuned, wei2022emergent,achiam2023gpt, anil2023palm,srivastava2022beyond}. The success of LLMs lies in their scale with billions of parameters and their pretraining on millions of tokens \citep{hoffmann2022training}. Nevertheless, deploying an LLM usually needs multiple GPUs and the huge amount of parameters induces long latencies for an LLM to complete its computation. These limitations have raised significant sustainability concerns~\citep{rillig2023risks}. In response, there is an ongoing quest to enhance the efficiency of LLMs, such as model distillation\citep{hinton2015distilling,gu2023minillm,jiao2019tinybert,wang2021want, wang2023distill}, model quantization\citep{yao2022zeroquant,bai2021binarybert,zafrir2019q8bert, xiao2023smoothquant, chee2024quip}, and model pruning\citep{lecun1989optimal,frantar2022optimal,hassibi1993optimal}.

Pruning reduces the size of an LLM by removing components while aiming to keep performance. 
Pruning methods can be categorized into structured pruning and unstructured pruning. Unstructured pruning, which removes connections between neurons, can often preserve performance effectively but requires specialized hardware to accelerate the pruned model.
In contrast, structured pruning targets the removal of entire neurons, attention heads, channels, and layers, offering intrinsic speedup without the need for specific hardware. However, some structured pruning methods impose specific constraints during pruning. For instance, several works~\citep{ashkboos2023slicegpt,frantar2023sparsegpt,ma2023llm,xia2023sheared} enforce the same sparsity across all layers, implicitly assuming equal importance among the layers---a premise that is not necessarily true. Besides, recent approaches~\citep{men2024shortgpt,gromov2024unreasonable} selectively remove transformer blocks, treating the attention layer and feed-forward network (FFN) within the same transformer block as a whole. This constraint presumes similar and interdependent importance between attention and FFNs. Although these constraints simplify the pruning problem, they introduce rigidity when searching for candidates, potentially limiting the flexibility needed to achieve the best pruning outcomes.

\begin{figure}[t]
\vspace{-12pt}
\centering
    \includegraphics[width=\columnwidth]{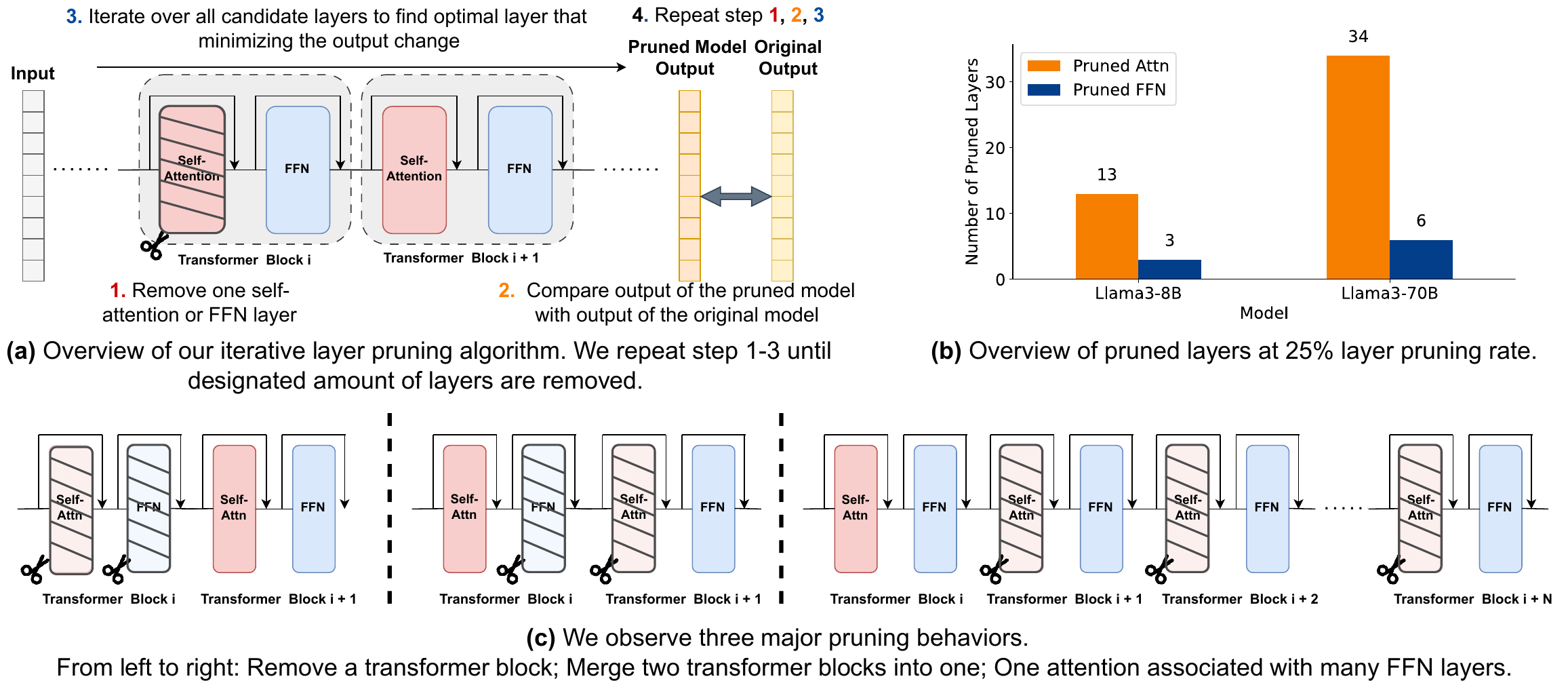}
    \caption{\textbf{(a)}: Overview of \proj{}. \proj{} iteratively examines candidate attention and FFN layers to find the next pruning target that minimizes output discrepancy compared to the original model. \textbf{(b)}: Overview of pruned layers in type. More attention layers are removed than FFN layers. \textbf{(c)}: Three major pruning behaviors we observed. Apart from pruning a transformer block or merging multiple transformer blocks to one transformer block through pruning, \proj{} tends to remove attention layers in consecutive transformer blocks.
}   
    \vspace{-5pt}
    \label{fig:method_overview}
\end{figure}

In this work, we introduce \proj{}, a flexible and effective layer pruning method that treats self-attention and FFN layers as separate pruning candidates. This approach offers a finer granularity compared to previous layer pruning methods. Another significant distinction lies in how we assess the impact of layers. Unlike previous layer pruning methods that measure the similarity between layer input and output, focusing on the local effects of the layers, our method evaluates the global impact by identifying layers \emph{whose removal causes minimal alteration to the model's output}. To identify which layers to prune, we employ an iterative algorithm (\cref{fig:method_overview} (a)). The algorithm not only evaluates individual layers but also considers the interdependencies among them, where the importance of remaining layers may shift following the removal of others. Evaluations on multiple tasks across various LLMs demonstrate that our pruning technique surpasses existing baselines and better preserves the capabilities of the original models. Furthermore, our method can serve as a mechanistic interpretation tool to study the importance of layers. Our analysis on pruned layers (\cref{fig:method_overview} (b)-(c) and \cref{sec:analyze_pruned_layers}) shows that self-attention layers located deeper in LLMs are more redundant than FFNs, which suggests a heterogeneous layer design for future LLMs. 

\textbf{Our contribution: } (i) We propose \proj{}, a novel layer pruning method that aims to reduce the computation of LLMs by treating self-attention and FFN layers as individual pruning candidates. (ii) We introduce a new formulation in model pruning aimed at minimizing the pruning effect on the model's output, measured by the shift in the predictive distribution. This formulation considers the characteristics of the entire model instead of only a target layer. Furthermore, it is task-agnostic. (iii) Compared with baseline approaches on various models and tasks, our model can effectively reduce the amount of computation of an LLM while better preserving its zero-shot ability on many tasks. (iv) We demonstrate the utility of our pruning method as a tool for mechanistic interpretability studies of LLMs. Our analysis of the pruned layers reveals that self-attention layers positioned later in LLMs are more redundant. Our observation suggests a heterogeneous layer design that could potentially enhance efficiency beyond the current homogeneous architectures.

\section{Related works}
\textbf{Unstructured pruning of LLMs: }The most primitive unstructured pruning is through 
magnitude-based removal of weights by setting weights with small values to zero, which performs poorly on LLMs~\citep{frantar2023sparsegpt}. One better way is through Optimal Brain Surgeon (OBS)~\citep{hassibi1993optimal,lecun1989optimal} that systematically removes weights that have the least impact on the loss function. However, OBS is impractical for large models such as LLM due to the complexity of calculating the Hessian. Optimal
Brain Compression \cite{frantar2022optimal} is a variant that decomposes the full-model pruning into per-layer unstructured pruning subproblems to reduce the computation of OBS and makes OBS applicable to LLMs. Furthermore, \citet{singh2020woodfisher} approximate Hessian matrix through block Fischer matrix. SparseGPT~\citep{frantar2023sparsegpt} formalizes the layer-wise unstructured pruning as a layer output reconstruction problem and solves it iteratively. \citet{sun2023simple} proposed to prune connections based on both weights and activations that achieved good performance without needing post-pruning reconstruction. 

\textbf{Structured pruning of LLMs: }
Structured pruning methods do not rely on special hardware realization and can intrinsically speed up a model. LayerDrop~\citep{fan2019reducing}, Block Pruning~\citep{francois2021block}, and ShearedLlama~\citep{xia2023sheared} introduce structured pruning as a regularization during pre-training or fine-tuning. Another line of research is to perform structured pruning after a model is already trained. SliceGPT~\citep{ashkboos2023slicegpt} removes rows and columns of a weight matrix, equivalent to reducing a layer's input/output dimension. LLM-Pruner~\citep{ma2023llm} removes parts of LLMs such as neurons, attention heads, or channels based on an importance score consisting of gradients of connected weights. Recently, more works have focused on reducing layers of LLMs. LaCo~\citep{yang2024laco} selects and merges multiple layers into one layer. ShortGPT~\citep{men2024shortgpt} removes decoder layers of LLMs based on cosine similarity between the input and output of decoder layers. \citet{gromov2024unreasonable} prunes a block of consecutive decoder layers according to angular distances between the block input and block output. Compared to other layer pruning methods, ours first proposes to look into decoder layers and treat self-attention layers and FFN layers separately as independent components to be pruned.

\section{Method}
\subsection{Preliminaries of LLMs}
Decoder-only LLMs generally comprise an embedding layer $E$, succeeded by $L$ transformer decoder layers $H_1, H_2, \ldots, H_L$, and conclude with a prediction head layer $C$. Each decoder layer $H_l$ incorporates an attention layer and an $FFN$ layer. Consider an input prompt $\xv \in |\vocab|^N$, where $\vocab$ represents the vocabulary and $N$ is the length of the prompt sequence. The LLM, denoted by $f$, processes the input as follows: Initially, $\xv$ is mapped into a hidden space as $\hv_0 = E(\xv) \in \mathbb{R}^{N \times d}$. Subsequently, the hidden state $\hv_0$ is passed through the decoder layers:
\begin{equation}
    \begin{aligned}
        \hv_l' &= \Attn(\hv_{l-1}) + \hv_{l-1} \quad \text{for } l = 1, 2, \ldots, L, \\
        \hv_l &= \FFN(\hv_l') + \hv_l' \quad \text{for } l = 1, 2, \ldots, L.\\ 
    \end{aligned}
\end{equation}
This formulation intentionally omits positional embeddings and normalization steps in $\Attn$ and the normalization in $\FFN$ to enhance clarity. Next, the head layer $C$ predicts the logits $C(\hv_L) = [\zv^{(1)}, \zv^{(2)}, \ldots, \zv^{(N)}]$, where $\zv^{(i)} \in \mathbb{R}^{|\vocab|}$ represents the predicted logits for the $(i+1)$-th output token. Although next-word prediction can be implemented in an auto-regressive manner, this mode of prediction is not employed during model pruning. Instead, we concentrate on the non-auto-regressive prediction approach.

\subsection{Formulation of structured pruning for LLMs}
Structured pruning begins by dividing the parameters $\thetab$ of the decoder layers into different groups based on the model's structure, and then prunes these parameters group-wise. For example, parameters that are in the same row or column of a weight matrix, or associated with the same neuron in an LLM, can be grouped together. In this work, we group the decoder parameters by $\Attn$ or $\FFN$ layers.

When pruning the model, we selectively drop some layers from the $2L$ $\Attn$ or $\FFN$ layers. The pruning layer selection can be parametrized by an \emph{indicator vector} $\mv \in \{0, 1\}^{2 L}$, where a value of $1$ signifies that the parameters within the corresponding $\Attn$ or $\FFN$ layer should be dropped. Then, the layer pruning ratio $r$ can mathematically described as:
\begin{equation}
    \begin{aligned}
        r = (\mv^T \ones) / (2L) , \quad \text{where } \ones = [1, \ldots, 1]^T.
    \end{aligned}
\end{equation}

The primary objective of pruning, for a specific $r$, is to identify the layers whose removal minimally impacts the model performance. Although performance evaluations across different tasks utilize diverse metrics, these metrics are intrinsically dependent on the model’s output. Therefore, to minimize performance degradation, the output of the pruned model $f(\xv; \thetab, \mv) = [\ztv^{(1)}, \ldots, \ztv^{(N)}]$ should closely approximate the original model output $f(\xv; \thetab) = [\zv^{(1)}, \ldots, \zv^{(N)}]$. This leads to the following optimization formulation for an optimal pruning layer selection $\mv^*$:
\begin{equation}
    \label{eq:opt_formulation}
    \begin{aligned}
        \mv^* = \arg\min_{\mv} \quad & \E_{\xv\sim p(\xv)}\left[ \frac{1}{N} \sum_{i=1}^N q\left(\zv^{(i)}, \ztv^{(i)} \right)\right] \\
        \text{s.t.} \quad & \mv^T \ones = 2L \cdot r,
    \end{aligned}
\end{equation}
where $q: \mathbb{R}^{|\vocab|} \times \mathbb{R}^{|\vocab|} \rightarrow \mathbb{R}_{\geq 0}$ is a metric used to measure changes in the model output. We can perform Monte-Carlo simulation on $\xv$ to get an approximation of the expectation. Although the optimization problem in \cref{eq:opt_formulation} appears straightforward, solving it is computationally challenging. A brute-force search for the global optimum would have \emph{exponential} computational complexity, which is infeasible to compute for a large $L$. Furthermore, the optimization parameter is a binary vector, making the gradient-based optimization not applicable. We discuss an iterative algorithm in the subsequent section to address these challenges.

\subsection{Iterative search algorithm as an efficient and approximate solver}
In this work, we utilize an iterative search algorithm as an approximation method, which reduces the complexity to $O(L^2)$. This algorithm progressively ablates one layer at a time, selecting the layer whose removal least affects the model output. This procedure is iterated upon, layer by layer, until the desired layer pruning ratio $r$ is achieved. The specifics of this method are detailed in \cref{alg:iterative_search}.

\begin{algorithm}
\caption{Iterative layer pruning}\label{alg:iterative_search}
\begin{algorithmic}
    \State $\mv \gets [0, 0, \ldots, 0]$
    \While{$r$ is not reached:}
        \State $Q_{\min} \gets \infty$, and $l_{\min} \gets 0$  \Comment{Initialize the change at output and the best pruning layer.}
        \For{$l$ in $\argwhere(\mv_l \neq 1)$}  \Comment{Loop over the remaining layers.}
            \State $\mv_l \gets 1$  \Comment{Try pruning layer $l$; update $\mv$ temporarily.}
            \State $Q \gets \mathsf{EvalOutputChange}(f, \thetab, \mv)$  \Comment{The change at output after the removal of layer $l$.}
            \If{$ Q \leq Q_{\min}$}
                \State $Q_{\min} \gets Q$, and $l_{\min} \gets l$  \Comment{Update the best pruning layer to layer $l$.}
            \EndIf
            \State $\mv_l \gets 0$  \Comment{Reset $\mv$ to previous state; prepare for ablating the next layer.}
        \EndFor
        \State $\mv_{l_{\min}} \gets 1$ \Comment{Prune the optimal single layer at the current step.}
    \EndWhile
\end{algorithmic}
\end{algorithm}

The function $\mathsf{EvalOutputChange}(f, \thetab, \mv)$ evaluates the output change between the original model and the pruned model with $q\left(\zv^{(i)}, \ztv^{(i)} \right)$ in \cref{eq:opt_formulation}. We discuss the choice of metric functions for measuring the output change in detail in \cref{sec:metric_functions}. 
Algorithm~\ref{alg:iterative_search} provides an efficient approximation instead of solving the optimization problem exactly. Nevertheless, the algorithm can be time-consuming, as modern LLMs usually have many layers. For a single pruning step in Algorithm~\ref{alg:iterative_search}, we still need to perform many forward passes on all remaining layers to find the pruning target. To further reduce the runtime, we adapt a finding from prior work \cite{gromov2024unreasonable} and consider only a subset of the last $60\%$ layers when the layer pruning ratio does not exceed $40\%$. For layer pruning ratios larger than $40\%$, we consider all layers as pruning candidates. Subsequently, we always consider only $60\%$ of all layers at any pruning step, hence further reducing the runtime of our pruning method.

\subsection{Choices of metric functions} \label{sec:metric_functions}
To evaluate changes in the model output, we consider multiple distance metrics for $q(\cdot, \cdot)$ in \cref{eq:opt_formulation}. In this study, we explore three metrics: (1) Euclidean distance, (2) angular distance, and (3) statistical distance. For notation clarity, the superscript $(i)$ in the output logits is omitted in this discussion. The metrics are introduced as follows:

\textbf{Angular distance:} Angular distance measures the distance between two vectors in hypersphere coordinates. This metric is used in several prior works~\citep{men2024shortgpt,gromov2024unreasonable}. Specifically, it measures the difference in orientation of two vectors. Formally, we leverage the cosine-similarity to measure the angular distance: 
\begin{equation}
    \begin{aligned}
        q(\zv, \ztv) = \arccos\left( \frac{\zv^T \ztv}{\|\zv\|_2 \|\ztv\|_2} \right)
    \end{aligned}
\end{equation}

\textbf{Euclidean distance:} One drawback of the previously discussed angular distance is that it can consider two dissimilar entities as identical. If two vectors have different norms but the same orientation, their angular distance is zero. To improve the pruning performance, we can apply more strict distance metrics. Since we can consider the model output as vectors in high dimensional space, it is natural to measure the Euclidean distance between two outputs. Formally, we have: 
\begin{equation}
    \begin{aligned}
        q(\zv, \ztv) = \sqrt{\sum\nolimits_{j=1}^{|\vocab|} (\zv_j - \ztv_j)^2}
    \end{aligned}
\end{equation}

\textbf{Statistical distance:} Instead of treating model outputs as deterministic vectors, one alternative perspective is to consider them as random variables. Hence, we can apply statistical measures to them. This viewpoint is natural, as model outputs after the softmax operation can be seen as a categorical distribution representing the confidence of predicting each possible token as the next token. Formally, given that $\zv$ and $\ztv$ represent logits, we apply the softmax function to derive the predicted probability distributions $\sv = \softmax(\zv)$ and $\stv = \softmax(\ztv)$. We then use a statistical distance to quantify the discrepancy between these two distributions. There are many options for statistical distance measurement. In this work, we choose the Jensen-Shannon divergence, a variant of Kullback–Leibler divergence with symmetric property:
\begin{equation}
    \begin{aligned}
        q(\zv, \ztv) = \frac{1}{2} \left[ \KL\left(\sv \ || \ \frac{1}{2}(\sv + \stv) \right) + \KL\left(\stv \ || \ \frac{1}{2}(\sv + \stv) \right) \right]
    \end{aligned}
\end{equation}
where $\KL$ represents the Kullback–Leibler divergence, defined as $\KL(\uv, \vv) = \sum_j \uv_j \log(\uv_j / \vv_j)$ for discrete random variables, with vectors $\uv$ and $\vv$ representing discrete probability distributions.

\section{Experiments}\label{sec:exp}
\subsection{Experiment setup}\label{sec:setup}
\textbf{Models: }
We consider a wide range of models for pruning. Specifically, we choose some member models of Llama family\cite{touvron2023llama1,touvron2023llama} including Llama2-7B, Llama2-13B, Llama3-8B, and Llama3-70B. The rationale for selecting these models is to evaluate the pruning methods on models at different parameter levels and across different generations. In addition, we include Mixtral-8x7B~\cite{jiang2024mixtral} in our experiments to assess the pruning performance on Mixture-of-Experts (MoE) models~\cite{fedus2022switch}. Due to the page limit, we present the results of Llama2 models in \cref{app:llama2_7b} and \cref{app:llama2_13b}.

\textbf{Benchmarks: }
We conduct our experiment on two types of tasks: generative tasks and zero/few-shot tasks. For generative tasks, we evaluate the perplexity on WikiText2~\citep{merity2016pointer}.
For zero/few-shot tasks, we evaluate accuracy on commonsense reasoning
datasets: BoolQ~\citep{clark-etal-2019-boolq}, PIQA~\citep{Bisk2020piqa}, HellaSwag~\citep{zellers2019hellaswag}, WinoGrande~\citep{ai2:winogrande}, ARC-easy/challenge~\citep{allenai:arc},OpenbookQA~\citep{OpenBookQA2018} and MMLU~\citep{hendrycks2021ethics,hendrycks2021measuring}. We use the LM-Evaluate-Harness framework~\citep{eval-harness} to ensure a transparent and reproducible evaluation by following exactly the prompts and evaluation metrics in the framework. More details are in \cref{app:detal_exp}. 

\textbf{Baselines: }
We include other four structured pruning methods in our evaluation: LLM-Pruner~\cite{ma2023llm}, SliceGPT~\cite{ashkboos2023slicegpt}, ShortGPT~\cite{men2024shortgpt}, and the pruning method discussed in \citet{gromov2024unreasonable}, which we refer it as \emph{DeeperLayers} in the following text. For ShortGPT and DeeperLayers, the implementation is not publicly available. Hence, we reproduce their methods according to their manuscripts. Our reproduced version is available in our codebase. For SliceGPT and LLM-Pruner, we use the official implementation. However, SliceGPT and LLM-Pruner are not designed to prune layers. Hence, we set the sparsity for these methods to be identical to our layer pruning methods. It is worth noting that LLM-Pruner currently does not support group-query attention (GQA)~\citep{ainslie2023gqa}, so we only apply LLM-Pruner on Llama2 models.

\textbf{Pruning settings: }
We randomly selected 10 samples from the Wikitext2 dataset for running our iterative layer pruning algorithm. To compare the actual performance of the pruning methods, we \textbf{did not conduct post-pruning training} to heal or reconstruct the pruned LLMs. In the following text, we refer to the pruning results associated with angular distance as Acos, those associated with Euclidean distance as Norm, and those associated with statistical distance as JS. For Llama3-8B, the pruning algorithm was executed on a single NVIDIA A100-SMX-80GB GPU. For Llama3-70B and Mixtral-8x7B, the pruning algorithm was run on two NVIDIA A100-SMX-80GB GPUs.

\subsection{Main result}
\cref{tab:llama3_8b_25_ratio} and \cref{tab:llama3_70b_25_ratio} show the performance of Llama3-8B and Llama3-70B when $25\%$ of the layers are pruned using different pruning methods. On Llama3-8B, \proj{} outperforms all other baselines and achieves $60\%$ mean accuracy on $8$ different reasoning tasks, which is around $20\%$ higher than the best prior work method. Furthermore, with $25\%$ of the layers removed, our pruned model still retains $88\%$ of the dense model's performance. On Llama3-70B, \proj{} retains $98\%$ of the original performance. We also evaluate the language modeling ability of pruned models with the WikiText2 dataset. According to the perplexity results, \proj{} better preserves the text generation ability than other baselines. Based on the performance drop on Llama3-70B and Llama3-8B, it is worth mentioning that a larger model is also an easier pruning target. This suggests more redundancy on larger models despite their superior performance compared to smaller models.

\begin{table}[t]
\centering
\caption{Performance on Llama3-8B at $25\%$ layer pruning ratio. ``Average'' shows the mean accuracy across 8 reasoning tasks. Acos, Norm, and JS are our approaches with different measures.  Our pruning method outperforms baselines on all evaluated tasks. The $^*$ symbol denotes that we set sparsity to be 25\% instead of layer pruning ratio. Evaluation: Wikitext is ppl, other tasks are accuracy. }
\resizebox{\linewidth}{!}{
\begin{tabular}{c |c |c | c c c c c c c c c}
\toprule
Model & Speedup & Wikitext & BoolQ & ARC-C & ARC-E & WG & HS & PIQA & OBQA & MMLU & Average \\ \hline
Original & - & 11.31 & 80.8 & 53.1 & 76.9 & 71.7 & 68.8 & 81.3 & 45.0 & 66.2 & 68.0 \\ \hline
ShortGPT & 1.31$\times$ & $10^3$ & 37.8 & 31.2 & 37.7 & 53.4 & 37.9 & 62.6 & 28.2 & 35.7 & 40.6 \\
DeeperLayers & 1.31$\times$ & $10^3$ & 37.8 & 31.2 & 37.7 & 53.4 & 37.9 & 62.6 & 28.2 & 35.7 & 40.6 \\
SliceGPT$^{*}$ & 1.29$\times$ & 45.65 & 37.6 & 28.8 & 45.8 & 59.5 & 51.9 & 58.5 & 33.6 & 24.2 & 42.5 \\ 
\hline
Acos & 1.30$\times$ & 23.57 & 72.7 & 35.8 & 62.2 & 65.9 & 60.1 & \textbf{74.8} & 37.8 & 41.2 & 56.3 \\
Norm & 1.30$\times$ & 38.14 & \textbf{78.1} & 37.5 & 63.6 & \textbf{67.6} & 60.7 & 73.8 & \textbf{39.0} & 46.7 & 58.4 \\
JS & 1.30$\times$ & \textbf{23.06} & 76.0 & \textbf{38.2} & \textbf{65.6} & 65.2 & \textbf{60.5} & 74.2 & 38.2 & \textbf{62.8} & \textbf{60.1} \\ 
\bottomrule
\end{tabular}
}
\label{tab:llama3_8b_25_ratio}
\end{table}

\begin{table}[t]
\centering
\caption{Performance on Llama3-70B at $25\%$ layer pruning ratio.}
\resizebox{\linewidth}{!}{
\begin{tabular}{c |c |c |c c c c c c c c c}
\toprule
Model & Speedup & Wikitext & BoolQ & ARC-C & ARC-E & WG & HS & PIQA & OBQA & MMLU & Average \\ \hline
Original & - & 8.22 & 84.8 & 65.0 & 85.3 & 79.8 & 75.4 & 85.3 & 48.6 & 79.8 & 75.5 \\ \hline
ShortGPT & 1.31$\times$ & 15.66 & 83.9 & 56.7 & 79.1 & 77.3 & 69.3 & 77.9 & 42.2 & \textbf{79.9} & 71.0 \\
DeeperLayers & 1.31$\times$ & 19.34 & 84.2 & 54.7 & 79.2 & 78.3 & 69.6 & 79.1 & 41.4 & 79.2 & 70.7 \\
SliceGPT$^{*}$ & 1.30$\times$ & 20.75 & 48.9 & 46.7 & 71.3 & 69.6 & 61.7 & 69.0 & 39.8 & 54.2 & 57.6 \\ 
\hline
Acos & 1.30$\times$ & 11.65 & \textbf{85.1} & 60.6 & 82.6 & 77.9 & 72.3 & 82.0 & \textbf{47.6} & 78.6 & 73.3 \\
Norm & 1.30$\times$ & 12.35 & 85.1 & 61.7 & 82.8 & 77.6 & 73.3 & 83.2 & 47.6 & 78.3 & 73.7 \\
JS & 1.30$\times$ & \textbf{11.55} & 84.3 & \textbf{62.4} & \textbf{83.7} & \textbf{79.3} & \textbf{73.6} & \textbf{84.0} & 46.8 & 79.0 & \textbf{74.1} \\ 
\bottomrule
\end{tabular}
}
\label{tab:llama3_70b_25_ratio}
\end{table}

\begin{table}[b]
\centering
\caption{Model statistics of a pruned Llama3-70B model using \proj{} (JS).}
\begin{tabular}{c c c c c}
\toprule
Layer pruning ratio & MACs & Time (per sample) & \# parameters & Memory \\ \hline
0\% & 734690 GMacs & 9.50 s & 70.5 B & 134 GB \\ 
25\% & 587756 GMacs & 7.42 s & 61.1 B & 116 GB \\ 
50\% & 386504 GMacs & 5.09 s & 39.6 B & 76 GB \\ 
75\% & 202535 GMacs & 2.39 s & 22.0 B & 43 GB \\ 
\bottomrule
\end{tabular}
\label{tab:model_statistics}
\end{table}

\begin{figure}[t]
\centering
\begin{subfigure}[b]{0.3\textwidth}
    \includegraphics[width=\columnwidth]{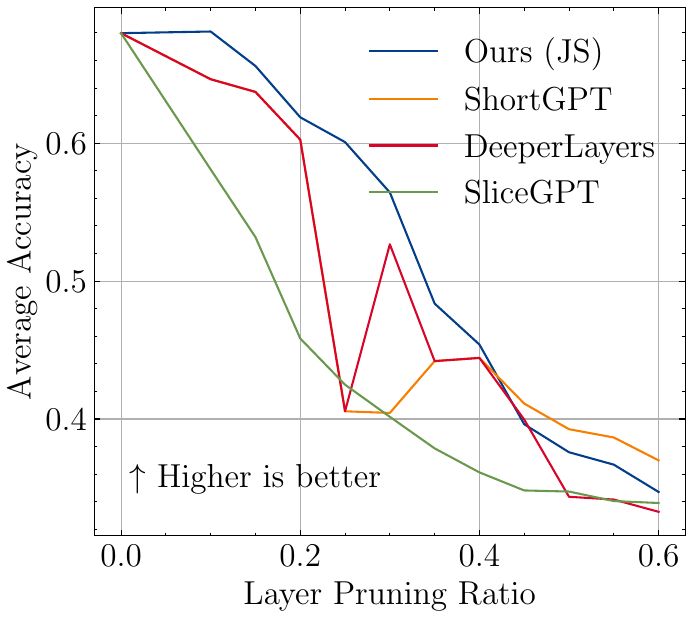}
    \caption{Llama3-8B}
    \label{fig:acc_diff_ratios_llama3_8b}
\end{subfigure}
\begin{subfigure}[b]{0.3\textwidth}
    \includegraphics[width=\columnwidth]{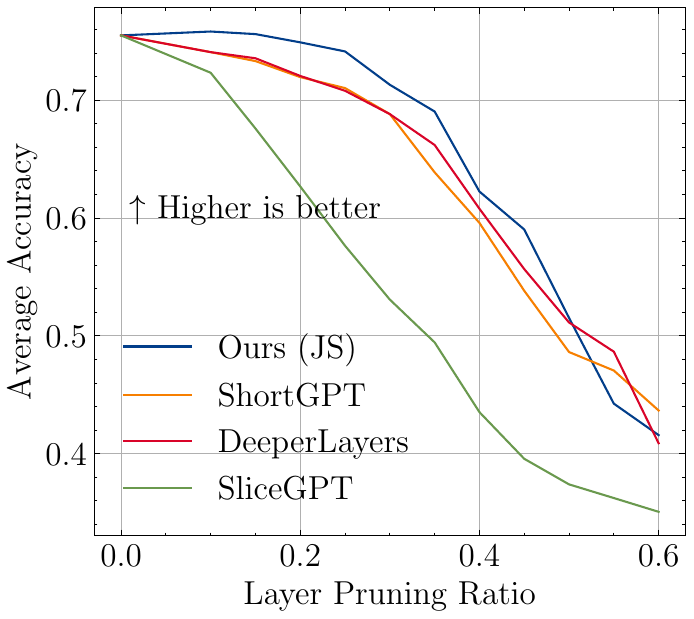}
    \caption{Llama3-70B}
    \label{fig::acc_diff_ratios_llama3_70b}
\end{subfigure}
\begin{subfigure}[b]{0.3\textwidth}
    \includegraphics[width=\columnwidth]{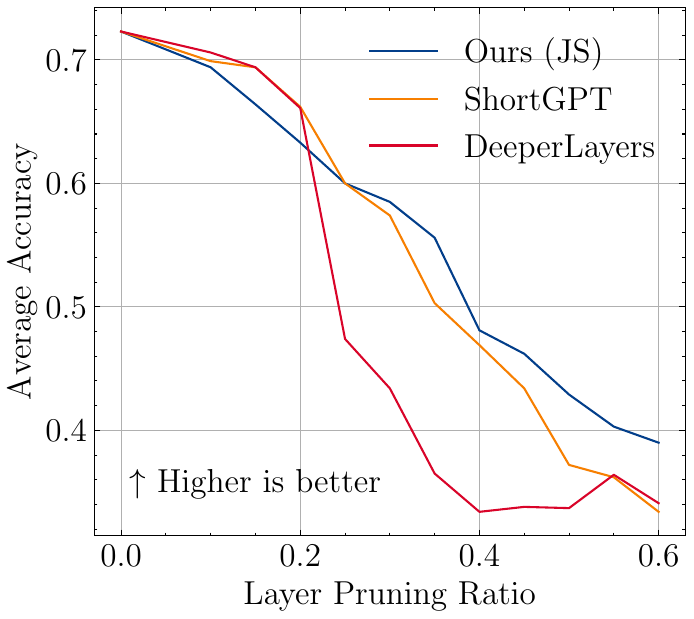}
    \caption{Mixtral-8x7B}
    \label{fig::acc_diff_ratios_mixtral_8x7b}
\end{subfigure}
\caption{Average zero/few-shot performance at different layer pruning ratios. We omit SliceGPT in (c) because it does not support MoE.}
\label{fig:acc_diff_ratios}
\end{figure}

\begin{figure}[t]
\centering
\begin{subfigure}[b]{0.3\textwidth}
    \includegraphics[width=\columnwidth]{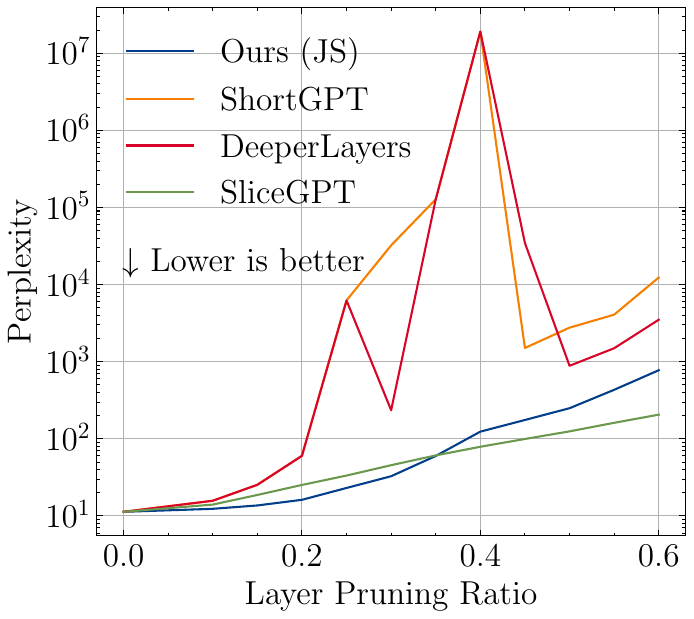}
    \caption{Llama3-8B}
    \label{fig:ppl_diff_ratios_llama3_8b}
\end{subfigure}
\begin{subfigure}[b]{0.3\textwidth}
    \includegraphics[width=\columnwidth]{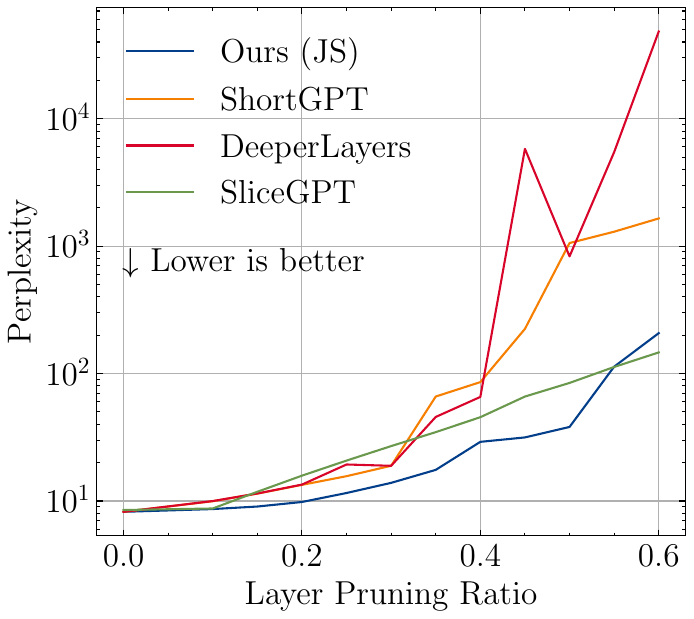}
    \caption{Llama3-70B}
    \label{fig:ppl_diff_ratios_llama3_70b}
\end{subfigure}
\begin{subfigure}[b]{0.3\textwidth}
    \includegraphics[width=\columnwidth]{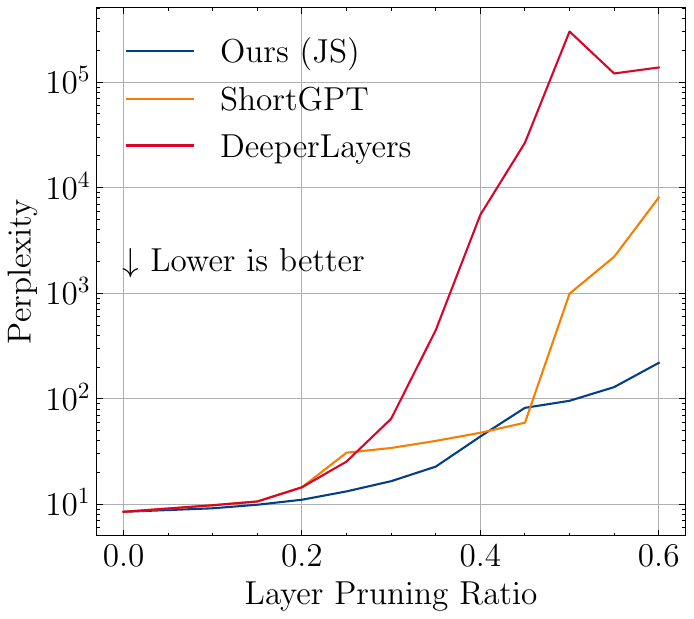}
    \caption{Mixtral-8x7B}
    \label{fig:ppl_diff_ratios_mixtral_8x7b}
\end{subfigure}
\caption{Perplexity (with a logarithmic scale) on WikiText2 at varying layer pruning ratios. Compared to ShortGPT and DeeperLayers, our method better preserves the language modeling capabilities.}
\label{fig:ppl_diff_ratios}
\end{figure}

To gauge the performance of pruning methods at different layer pruning ratios, we summarize our results in \cref{fig:acc_diff_ratios} and \cref{fig:ppl_diff_ratios}. Here, we only present our results with JS divergence. From \cref{fig:acc_diff_ratios} we see that our method performs better on zero-shot and few-shot reasoning tasks in the Llama3 model family. For Llama3-8B, our method still retains the model performance at $25\%$ layer pruning ratio, while other baselines have already collapsed. For language generation tasks shown in \cref{fig:ppl_diff_ratios}, our model performs substantially better than other baselines and can have more than $10$ times smaller perplexity on the WikiText2 generation task. In contrast, other methods like ShortGPT exhibit a stronger performance degradation on language generation tasks than on zero/few-shot tasks. Additionally, we demonstrate text examples generated by the \proj{}-pruned models in \cref{app:text_generation}.

\begin{table}[b]
\centering
\caption{Time and memory consumption for running different pruning methods on Llama2-7B. We use the Llama2-7B model such that we can include LLM-Pruner in our comparison. }
\begin{tabular}{c c c c c c}
\toprule
  & SliceGPT & LLM-Pruner & ShortGPT & DeeperLayers & \proj{} \\ \hline
Time consumption & 4 min & 1 min & 1 min & 1 min & 27 min\\ 
Memory usage & 25.2 GB & 40.71 GB & 27.15 GB & 27.15 GB & 27.19 GB \\ 
\bottomrule
\end{tabular}
\label{tab:prune_resource_comparison}
\end{table}

In addition, Table~\ref{tab:model_statistics} shows the computation and memory reduction of a \proj{}-pruned model. For memory measurements, we consider only the memory required to load the model. We measure MACs and runtime under an 8k context length, as the 8k context length is applied for pretraining. At $25\%$ layer pruning ratio, our methods can effectively reduce the amount of computation by around $25\%$, which is calculated in MACs. We observe that the runtime reduction is proportional to the MAC reduction as expected. At the $25\%$ layer pruning ratio, the memory usage of the pruned model is close to the memory usage of the original model. This is because our pruning methods choose to prune self-attention layers at early pruning iterations (details of pruned layers are presented in Section~\ref{sec:analyze_pruned_layers}). Since GQA in Llama3-8B is memory efficient, removing them can effectively reduce the runtime but barely reduces the memory usage. At larger pruning ratios, our method removes more FFN layers, hence resulting in more significant memory reduction. 

In Table~\ref{tab:prune_resource_comparison}, we compare the runtime and memory usage of our method with other baselines. Unlike the LLM-Pruner, our method does not require gradient information during pruning and relies solely on forward passes. This significantly reduces the GPU memory requirements, making our method's memory usage comparable to other layer pruning methods that also operate on forward passes alone. Conversely, methods that utilize gradient information consume more GPU memory because they must store activations during forward passes. This requirement constrains their scalability and limits their usage under low-resource conditions.
Regarding runtime, our method is lengthier because it iterates over all remaining layers before selecting the next layer to prune. However, we believe this increase in runtime is justifiable since the pruning process is a one-time requirement. The performance of the pruned model, which is a critical metric for any pruning method, outweighs the longer runtime.

\begin{figure}[t]
    \centering
    \includegraphics[width=\textwidth]{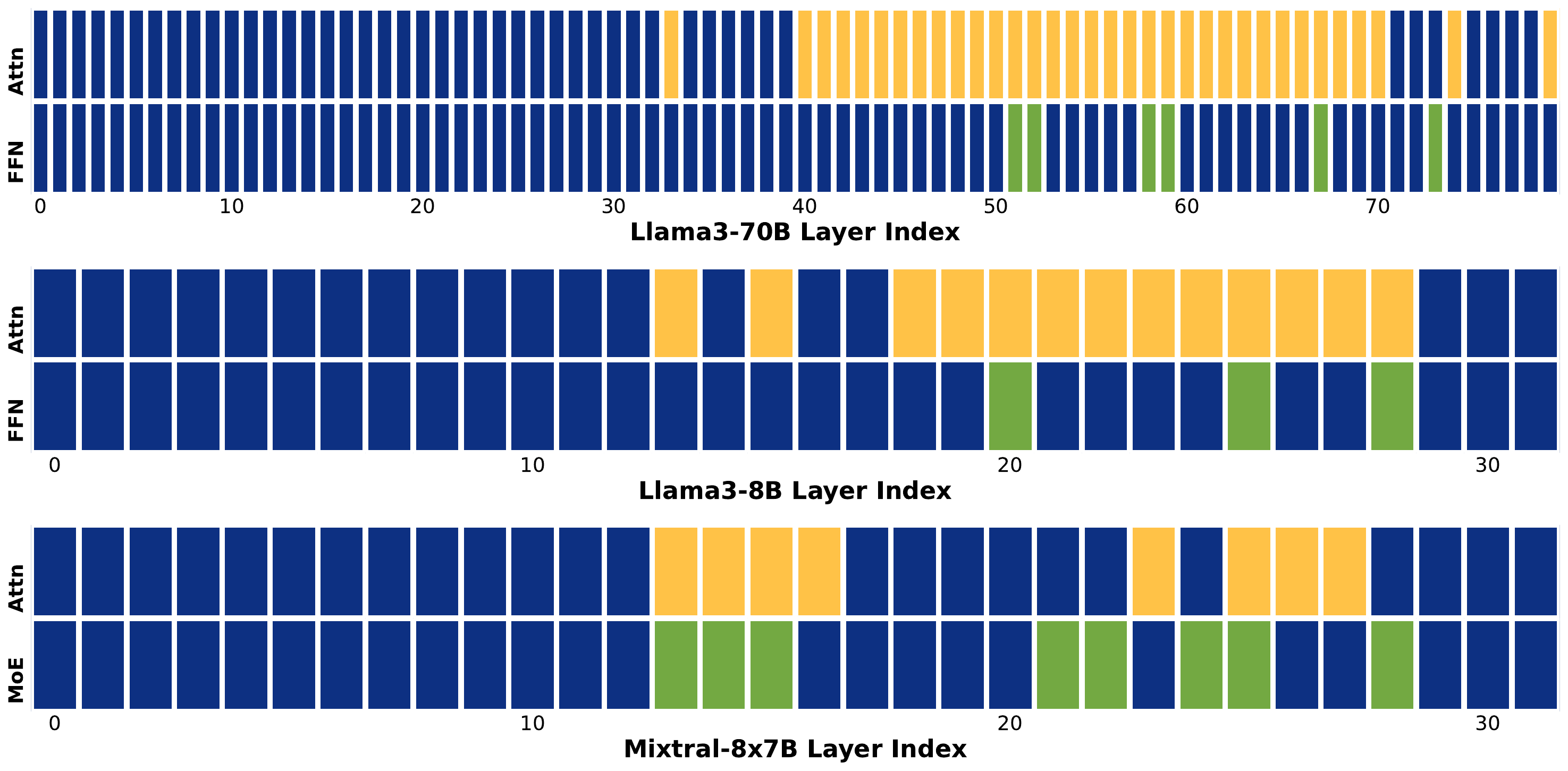}
    \caption{Visualization of pruned layers at $25\%$ layer pruning ratio for Llama3-70B (top, with $1.9\%$ performance drop), Llama3-8B (middle, with $11.6\%$ performance drop), and Mixtral-8x7B (bottom, with $17.0\%$ performance drop) using $\proj$. \textcolor{dropattncolor}{\rule{10pt}{6pt}} indicates pruned self-attention layers, \textcolor{dropffncolor}{\rule{10pt}{6pt}} indicates pruned FFN layers, and \textcolor{keepcolor}{\rule{10pt}{6pt}} indicates remaining layers. Notably, consecutive self-attention layers are removed, resulting in a heterogeneous structure where multiple FFNs process the output of one attention layer. More discussion in Section~\ref{sec:analyze_pruned_layers}. }
    \label{fig:pruned_layers_overview}
\end{figure}

\subsection{Analysis of pruned layers}\label{sec:analyze_pruned_layers}
Our layer pruning method can also work as an effective tool to study the mechanistic interpretability of LLMs. 
In this section, we present the results of pruned layers and study the layer importance of various LLM models based on the pruning results. 

\cref{fig:pruned_layers_overview} shows the visualization of pruned layers in Llama3-8B and Llama3-70B models. At $25\%$ layer pruning ratio, our pruning methods choose to mainly remove self-attention layers in the model. At this layer pruning ratio, pruned models exhibit minimal performance degradations, suggesting that these attention layers are unimportant. Moreover, the pruned self-attention layers are usually in consecutive transformer blocks. For instance, self-attention layers from index $40$ to $70$ are completely removed in Llama3-70B, and self-attention layers from index $18$ to $28$ are completely removed in Llama3-8B. Subsequently, we have a new structure at deeper layers, in which one self-attention layer is followed by multiple FFN layers. Surprisingly, on pruned Llama3-70B, there is even one attention layer followed by more than $20$ FFN layers. 
Based on these observations, we hypothesize that at later parts of LLMs, multiple consecutive FFN layers work in cooperation to process the output from one attention layer. This observation suggests that the current transformer architecture may not be the optimal structure for LLMs. Instead, non-uniform transformer structures with more FFNs at the later stage can be potentially more parameter-efficient and may be a better design option for future LLMs. 

We further inspect the pruning result on Mixtral-8x7B, an LLM equipped with Mixture-of-Experts (MoE) layers. \cref{fig:pruned_layers_overview} also shows pruned layers on Mixtral-8x7B. Intriguingly, the pruning selection is more balanced between attention and MoE layers than on Llama models. Though MoE layers are sparse structures, their representation capacity is larger as they contain multiple FFNs. Hence, we conjecture that the Mixtral-8x7B model is learned not to use multiple MoE layers to process for one attention layer. Our result suggests that MoE models may have less layer redundancy than conventional transformer models by design.    

Lastly, we also observe the removal and merging of transformer blocks in all pruned models. From the pruning result, we can see cases where one entire transformer block is pruned, as observed for instance on Llama3-70B at layer $51$, and on Mixtral-8x7B at layer $13$. In addition, multiple transformer blocks can be merged together. As shown in Mixtral-8x7B, layer $22$ and $23$ are merged into one transformer block by removing the MoE layer and the subsequent attention layer. 
\subsection{Ablation study}
\begin{figure}[t]
    \centering
    \includegraphics[width=\textwidth]{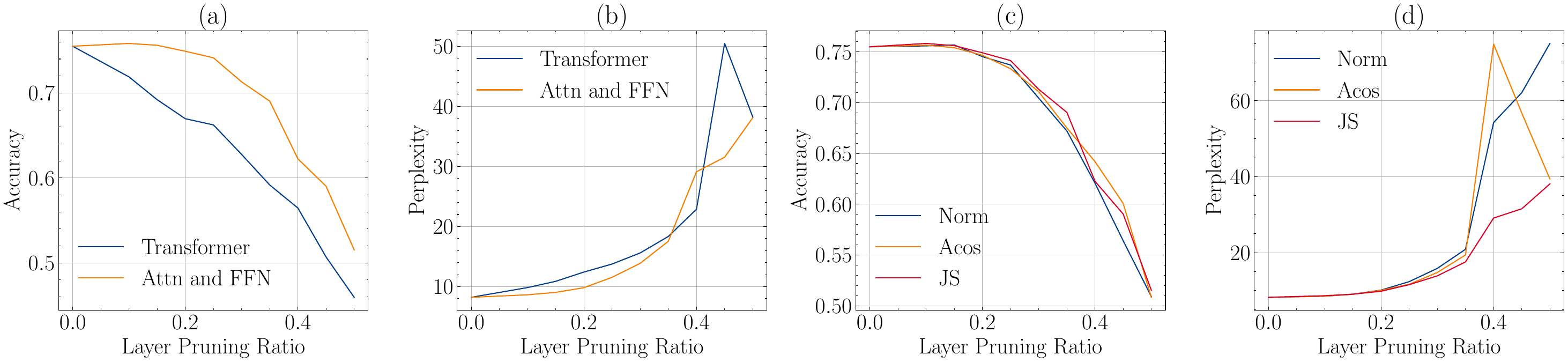}
    \caption{Ablation study on Llama3-70B. \textbf{(a)} and \textbf{(b)}: Pruning transformer blocks vs. pruning attention and FFN layers separately. \textbf{(c)} and \textbf{(d)}: Comparison of three distance metrics in \proj{}. }
    \label{fig:ablation_study}
\end{figure}

In this section, we verify the design choices of $\proj$ and examine their significance through two experiments. First, we evaluate the option of pruning transformer blocks instead of attention and FFN layers. In addition, we compare different metrics for distance measures. 
~\cref{fig:ablation_study} (a) and (b) compare the pruning performance when pruning candidates are different. The performance is measured in two setups: (i) considering attention and FFN layers as separate pruning candidates; or (ii) only pruning transformer blocks. The results on zero/few-shot tasks demonstrate the advantage of pruning attention and FFN layers compared to transformer blocks, highlighting the importance of pruning at a finer level.
\cref{fig:ablation_study} (c) and (d) show the pruning performance using three different distance measures: norm, arccos, and JS divergence. On zero/few-shot tasks, all three measures have similar performance. On perplexity results, JS divergence outperforms norm and arccos measures. Therefore, we recommend using JS divergence or other probabilistic measures as pruning metrics.

\section{Limitation}\label{sec:limitation}
In this work, we demonstrate that we can remove some layers while effectively retaining the performance of a LLM. 
Although our layer pruning method has notable performance improvements compared to other baselines, several limitations warrant further investigation. 
Our study mainly focused on three distance metrics. There might exist distance metrics that can yield better pruning results. Additionally, despite the low computational complexity of our iterative pruning method, it only provides a local optimal solution to the optimization problem. Advanced combinatorial optimization methods may identify better selections of pruning layers.

\section{Conclusion and future works}
In this work, we propose $\proj$, an effective layer pruning method with a finer scope that prunes attention and FFN layers. $\proj$ performs task-agnostic layer removal and can work under limited data and computational resources. We evaluate the efficacy of LLM-Pruner on three sets of distinct models, Llama2 family, Llama3 family, and Mixtral8x7b. Compared to other baselines, $\proj$ leads to better models that in general preserve $90\%$ of their performance with around $30\%$ layers removed, without fine-tuning or post-pruning reconstruction. Furthermore, our pruning results demonstrate that self-attention layers in LLMs are highly redundant in certain model architectures, which provides new insights into the inner mechanism of LLMs. These findings carry significant implications for the design of future LLMs.

Pruning methods that remove neurons and structures within layers can be combined with our method to achieve non-uniform pruning at the layer level, which can potentially improve the pruning performance. We hope that our research advances the understanding of LLMs pruning and sets the stage for further developments in efficient LLMs and their interpretability.
\newpage

\bibliographystyle{plainnat}
\bibliography{references}

\newpage
\appendix
\section{Broader impact}\label{app:broader_impact}
The method we proposed in this paper, \proj{}, is a layer pruning method that aims to prune large language models (LLMs). The development and implementation of methods to prune LLMs hold significant potential for both positive and negative impacts across various domains. Currently, LLMs are trained in computing clusters available only in certain countries and deployed in workstations equipped with costly GPUs. In a nutshell, they are not accessible to the public and people with limited computational resources. By reducing the size and computational requirements of LLMs, we can make LLMs, which have shown their capability and are publically considered powerful tools in many regards, more accessible to a wider range of users and applications. Potential use cases include LLMs on laptops, mobile devices, or IoT devices. This increased accessibility can foster innovation in many fields such as education, healthcare, and natural language processing by enabling smaller organizations and researchers with limited resources to leverage advanced AI technologies. In addition, increased accessibility also helps to avoid AI monopoly in the future.

Pruning LLMs can also contribute to environmental sustainability by decreasing the energy consumption and carbon footprint associated with training and deploying these models. As AI becomes increasingly integrated into everyday applications, the environmental impact of large-scale computations is a growing concern and cannot be neglected. Efficiently pruned models can help mitigate this issue by requiring fewer resources to operate effectively for both training and inference.

However, the broader adoption of pruned models must be approached with caution. Reducing model size can potentially lead to a loss of nuance and accuracy in language understanding and generation, which could have unintended consequences in critical applications such as medical diagnosis or legal analysis. Furthermore, the democratization of powerful language models, while generally positive, raises ethical considerations regarding the potential misuse of these technologies for generating misleading or harmful content. Hence, further research and collaboration across disciplines will be crucial in ensuring that the deployment of pruned LLMs is both responsible and beneficial to society.

\section{Detailed experiment settings}\label{app:detal_exp}
We used eight datasets to evaluate the language models: Wikitext2, BoolQ, ARC, WinoGrande, HellaSwag, PIQA, OpenbookQA and MMLU.
Other than WikiText, which evaluates the text generation performance by measuring perplexity, a metric that evaluates how similar is the return to a real text for a fixed-lengthen response. All other tasks are multiple-choice QA problems. 
For MMLU, we evaluated the model in a 5-shot setting, and for other multiple-choice tasks, we evaluated it in a zero-shot setting. Following other evaluation approaches, all zero/few-shot tasks are evaluated by mean accuracy. We compute an accumulated accuracy across all tokens in the target answer and then divide the result by the token length of the target answer.

\paragraph{Wikitext2}
The WikiText language modeling dataset consists of over 100 million tokens that have been extracted from the set of verified Good and Featured articles on Wikipedia. The task is to measure the perplexity on the Wikitext dataset, via rolling loglikelihoods. A lower perplexity means better text generation performance.
\paragraph{BoolQ}
The BoolQ dataset contains 15942 examples of yes/no questions. The questions are generated naturally – they occur in an unprompted and unrestricted environment. Each example is a triplet (questions, passages, answers), with a title as an optional context.
\paragraph{ARC}
The ARC dataset includes 7,787 science exam questions selected from various sources, including questions provided by AI2 affiliates under license.
The exam questions are text-only, English language questions that span several grade levels.
Questions are structured with multiple-choice answers (usually four options).
There are two sets of questions: the Challenge Set and the Easy Set. The Challenge Set has 2,590 "hard" questions (those that cannot be answered correctly by both retrieval and co-occurrence methods). The Easy Set has  5,197 questions.
\paragraph{WinoGrande}
WinoGrande consists of 44k fill-in-the-blank type problems with binary options.
It requires common sense reasoning to choose the right option for a given sentence.
\paragraph{HellaSwag}
HellaSwag is a challenge dataset for evaluating commonsense Natural Language Inference that is especially hard for models, though its questions are trivial for humans (>95\% accuracy). The questions are four-way multiple-choice problems.
\paragraph{PIQA}
PIQA (Physical Interaction: Question Answering) is a dataset for commonsense reasoning, designed to investigate the physical knowledge of the models. The underlying task is a binary choice question answering task.

\paragraph{OpenbookQA}
The OpenbookQA is based on open-book exams and is a question-answering dataset that assesses human understanding of subjects through question-answering. There are 5,957 multiple-choice questions addressing elementary science topics, which assess a candidate's understanding of 1,326 core science facts and the application of these facts in novel circumstances.
\paragraph{MMLU}
The MMLU consists of a series of 15,908 multiple-choice questions covering 57 academic disciplines including mathematics, philosophy, law and medicine as well as many other areas of academic study.
The questions are in the form of four-way multiple-choice questions. We adopt the frequently-used 5-shot setting to provide four sample questions and answers from the dev subset of MMLU, then ask the model to answer the actual question with A,B,C, or D.

\section{More evaluation result on Mixtral-8x7B}\label{app:mixtral_8x7b}
In complement to the \cref{fig::acc_diff_ratios_mixtral_8x7b} and \cref{fig:ppl_diff_ratios_mixtral_8x7b} in the main text, we present a more detailed pruning result on Mixtral-8x7B with performance reports on each task. We exclude ShortGPT and LLM-Pruner for this model because their implementations do not support MoE layers.
The results highlight the robustness and efficiency of \proj{} across various benchmarks. Our experiments demonstrate that our proposed pruning method, \proj{}, has many applications and can be used on MoE models effectively. 

\begin{table}[t]
\centering
\caption{Performance on Mixtral-8x7B at $25\%$ layer pruning ratio.}
\resizebox{\linewidth}{!}{
\begin{tabular}{c |c |c c c c c c c c c}
\toprule
Model & Wikitext & BoolQ & ARC-C & ARC-E & WG & HS & PIQA & OBQA & MMLU & Average \\ \hline
\text{Original} & 8.50 & 84.5 & 59.7 & 83.2 & 75.6 & 72.8 & 83.9 & 46.6 & 72.2 & 72.3 \\ \hline
\text{ShortGPT} & 30.86 & \textbf{83.2} & 39.4 & 54.0 & \textbf{69.5} & 57.7 & 69.3 & 37.6 & \textbf{69.6} & 60.0 \\
\text{DeeperLayers} & 25.42 & 54.6 & 30.2 & 54.3 & 51.4 & 53.5 & 72.9 & 37.0 & 25.6 & 47.4 \\
\hline
\text{Acos} & 13.54 & 77.8 & 42.1 & 67.5 & 66.9 & 64.5 & 78.1 & 39.6 & 39.3 & 59.5 \\
\text{Norm} & 14.24 & 79.1 & \textbf{43.9} & \textbf{68.5} & 65.9 & \textbf{65.9} & 78.1 & 40.8 & 43.5 & \textbf{60.7} \\
\text{JS} & \textbf{13.26} & 77.3 & 42.9 & 68.2 & 66.3 & 64.8 & \textbf{78.4} & \textbf{41.0} & 40.7 & 60.0 \\ 
\bottomrule
\end{tabular}
}
\label{tab:mixtral8x7b_25_ratio}
\end{table}

\begin{table}[t]
\centering
\caption{Performance on Mixtral-8x7B at $40\%$ layer pruning ratio.}
\resizebox{\linewidth}{!}{
\begin{tabular}{c |c |c c c c c c c c c}
\toprule
Model & Wikitext & BoolQ & ARC-C & ARC-E & WG & HS & PIQA & OBQA & MMLU & Average \\ \hline
\text{Original} & 8.50 & 84.5 & 59.7 & 83.2 & 75.6 & 72.8 & 83.9 & 46.6 & 72.2 & 72.3 \\ \hline
\text{ShortGPT} & 47.64 & \textbf{64.7} & 31.9 & \textbf{49.6} & 52.8 & 50.2 & 66.5 & 33.2 & 26.3 & 46.9 \\
\text{DeeperLayers} & 5335.09 & 44.2 & 27.5 & 25.5 & 49.1 & 26.8 & 48.2 & 24.6 & 23.2 & 33.4 \\
\hline
\text{Acos} & \textbf{43.31} & 58.9 & 31.9 & 44.5 & 57.5 & 54.0 & 64.3 & 29.4 & 24.7 & 45.6 \\
\text{Norm} & 55.77 & 40.2 & \textbf{35.2} & 48.5 & \textbf{60.3} & \textbf{57.3} & \textbf{66.7} & \textbf{33.8} & 24.5 & 45.8 \\
\text{JS} & 43.93 & 62.5 & 34.4 & 47.4 & 58.9 & 56.0 & 66.3 & 32.6 & \textbf{26.8} & \textbf{48.1} \\ 
\bottomrule
\end{tabular}
}
\label{tab:mixtral8x7b_40_ratio}
\end{table}

\section{More evaluation results on the Llama2 model family}\label{app:llama2}
In this section, we present additional experimental results on Llama2 models. While the main text focuses on Llama3 models, we include Llama2 results to provide a comprehensive overview. We report results on Llama3 models in the main text because the conclusions drawn from Llama2 models are consistent with those from Llama3, and we believe that results on newer models are more relevant to our audience. However, Llama2 results are included for completeness and because they include results of LLM-Pruner, which does not apply to Llama3 models. We use Llama2-7B and Llama2-13B models as pruning targets. We do not include the Llama2-70B model, as this model requires more resources to run.

\subsection{More evaluation results on Llama2-7B}\label{app:llama2_7b}
In this section, we present the evaluation results for pruned Llama2-7B models. Our analysis includes the same performance metrics as shown in the main text. \cref{tab:llama2_7b_25_ratio} shows the performance of a pruned model with $25\%$ layer removal ratio. Additionally, \cref{tab:llama2_7b_25_ratio} shows the performance of a pruned model with $40\%$ layer removal ratio.

\begin{table}[h]
\centering
\caption{Performance on Llama2-7B at $25\%$ layer pruning ratio. Both SliceGPT and LLM-Pruner are applied with a desired $25\%$ sparsity ratio according to their implementation. }
\resizebox{\linewidth}{!}{
\begin{tabular}{c |c |c c c c c c c c c}
\toprule
Model & Wikitext & BoolQ & ARC-C & ARC-E & WG & HS & PIQA & OBQA & MMLU & Average \\ \hline
\text{Original} & 10.14 & 76.2 & 46.0 & 74.0 & 68.3 & 66.5 & 79.7 & 44.2 & 47.1 & 62.8 \\ \hline
\text{ShortGPT} & 127.89 & 62.5 & 31.0 & 41.7 & 60.8 & 44.0 & 60.1 & 35.0 & \textbf{44.4} & 47.4 \\
\text{DeeperLayers} & 35.36 & 49.2 & 35.0 & 50.5 & 64.0 & 54.7 & 67.9 & 37.6 & 44.4 & 50.4 \\
\text{SliceGPT$^{*}$} & 20.05 & 52.1 & 34.5 & 55.6 & 62.9 & 55.1 & 67.5 & 35.8 & 29.7 & 49.1 \\ 
\text{LLM-Pruner$^{*}$} & 20.82 & 62.4 & 37.2 & 62.0 & 62.2 & 60.1 & \textbf{75.7} & 38.6 & 25.3 & 52.9 \\ 
\hline
\text{Acos} & 20.49 & 73.2 & 37.7 & \textbf{62.8} & 64.1 & 60.1 & 73.9 & \textbf{40.2} & 32.3 & \textbf{55.5} \\
\text{Norm} & 21.73 & \textbf{74.5} & \textbf{38.4} & 61.5 & 63.7 & \textbf{60.3} & 73.6 & 38.8 & 31.8 & 55.3 \\
\text{JS} & \textbf{18.43} & 69.3 & 38.4 & 62.1 & \textbf{64.4} & 59.5 & 75.4 & 39.8 & 28.5 & 54.7 \\ 
\bottomrule
\end{tabular}
}
\label{tab:llama2_7b_25_ratio}
\end{table}

\begin{table}[h]
\centering
\caption{Performance on Llama2-7B at $40\%$ layer pruning ratio. Both SliceGPT and LLM-Pruner are applied with a desired $25\%$ sparsity ratio according to their implementation.}
\resizebox{\linewidth}{!}{
\begin{tabular}{c |c |c c c c c c c c c}
\toprule
Model & Wikitext & BoolQ & ARC-C & ARC-E & WG & HS & PIQA & OBQA & MMLU & Average \\ \hline
\text{Original} & 10.14 & 76.2 & 46.0 & 74.0 & 68.3 & 66.5 & 79.7 & 44.2 & 47.1 & 62.8 \\ \hline
\text{ShortGPT} & 1683.61 & 55.4 & 27.7 & 25.6 & 50.3 & 30.1 & 50.7 & 26.6 & \textbf{39.6} & 38.3 \\
\text{DeeperLayers} & 303.03 & 62.6 & 32.0 & 35.1 & 59.2 & 39.3 & 59.4 & 27.6 & 39.6 & 44.4 \\
\text{SliceGPT$^{*}$} & \textbf{46.63} & 37.5 & 27.3 & 43.5 & 57.9 & 43.6 & 58.5 & 28.4 & 24.4 & 40.1 \\ 
\text{LLM-Pruner$^{*}$} & 48.39 & 50.2 & 31.3 & \textbf{50.7} & 56.1 & \textbf{53.5} & \textbf{70.7} & \textbf{36.6} & 24.4 & 46.7 \\ 
\hline
\text{Acos} & 73.36 & 36.8 & \textbf{34.0} & 46.9 & \textbf{62.0} & 49.5 & 66.8 & 33.4 & 25.6 & 44.4 \\
\text{Norm} & 66.37 & 63.2 & 32.5 & 42.7 & 57.6 & 48.9 & 64.7 & 34.0 & 25.0 & 46.1 \\
\text{JS} &	61.49 & \textbf{65.6} & 33.9 & 49.3 & 60.3 & 48.1 & 64.5 & 34.6 & 25.3 & \textbf{47.7} \\ 
\bottomrule
\end{tabular}
}
\label{tab:llama2_7b_40_ratio}
\end{table}

\begin{figure}[h]
\centering
\begin{subfigure}[b]{0.4\textwidth}
    \includegraphics[width=\columnwidth]{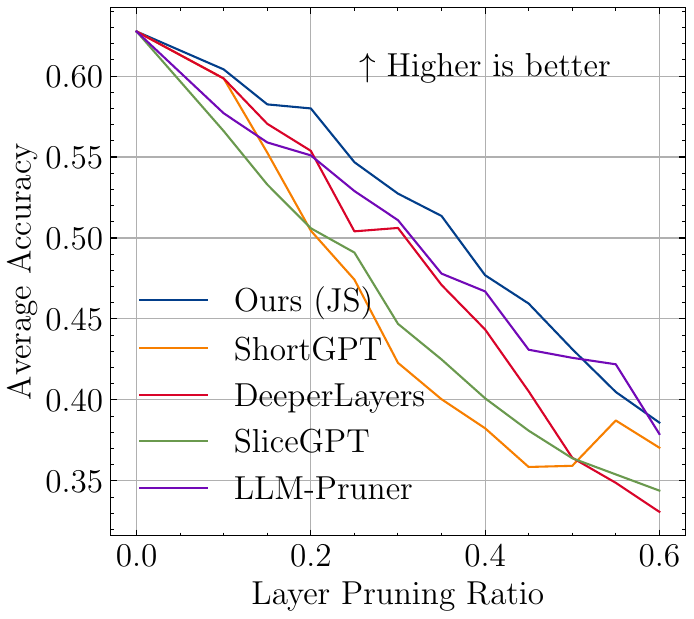}
    \caption{}
    \label{fig:acc_llama2_7b}
\end{subfigure}
\hspace{1cm}
\begin{subfigure}[b]{0.4\textwidth}
    \includegraphics[width=\columnwidth]{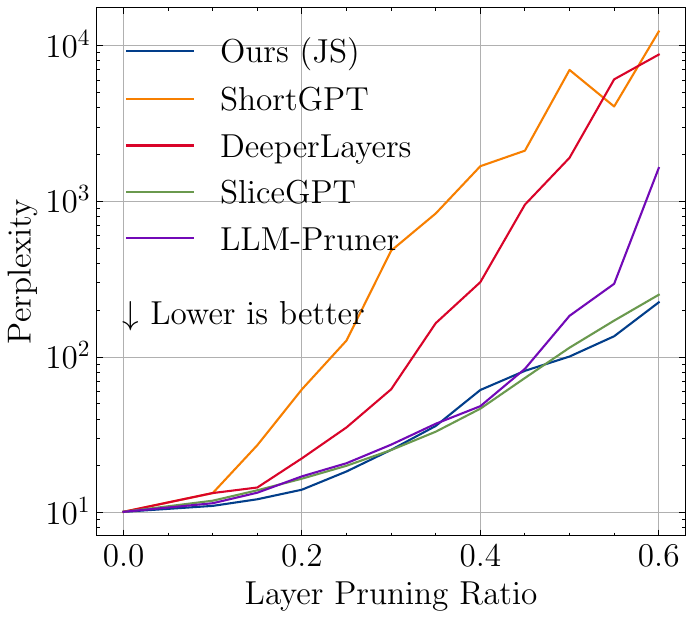}
    \caption{}
    \label{fig::ppl_llama2_7b}
\end{subfigure}
\caption{\textbf{(a)}: Average zero/few-shot performance at different layer pruning ratios on Llama2-7B. \textbf{(b)}: Perplexity (with a logarithmic scale) on WikiText2 at varying layer pruning ratios. \proj{} outperforms other methods including LLM-Pruner (not evaluated on Llama3 models) on both text generation and QA tasks. }
\label{fig:llama2_7b}
\end{figure}

To provide an overview how well pruning methods work in different cases. We include \cref{fig:llama2_7b} to show the performance change at various layer pruning ratios. According to the figure, \proj{} can in general better preserve the performance of pruned models. In addition, we include LLM-Pruner in all evaluations on Llama2-7B. LLM-Pruner performs better than other baseline methods but still less effective than \proj{}. Importantly, \proj{} can be applied to many models including models that do not exist yet, while LLM-Pruner needs to be adopted to those new structures.

\subsection{More evaluation results on Llama2-13B}\label{app:llama2_13b}
In this section, we present the evaluation results for pruned Llama2-13B models. 

\begin{table}[h]
\centering
\caption{Performance on Llama2-13B at $25\%$ layer pruning ratio. Both SliceGPT and LLM-Pruner are applied with a desired $25\%$ sparsity ratio according to their implementation.}
\resizebox{\linewidth}{!}{
\begin{tabular}{c |c |c c c c c c c c c}
\toprule
Model & Wikitext & BoolQ & ARC-C & ARC-E & WG & HS & PIQA & OBQA & MMLU & Average \\ \hline
\text{Original} &12.96 & 79.8 & 49.0 & 77.2 & 71.6 & 69.6 & 81.3 & 45.2 & 55.0 & 66.1 \\ \hline
\text{ShortGPT} & 39.15 & 62.7 & 41.9 & 60.1 & 70.5 & 60.6 & 73.1 & 40.8 & 48.6 & 57.3 \\
\text{DeeperLayers} & 17.90 & 57.9 & 43.8 & 64.6 & 69.5 & 61.3 & 73.9 & 39.0 & 54.4 & 58.1 \\
\text{SliceGPT$^{*}$} & 16.75 & 40.4 & 40.2 & 61.5 & 67.0 & 59.4 & 69.6 & 40.8 & 39.7 & 52.3 \\ 
\text{LLM-Pruner$^{*}$} &\textbf{14.84} & 66.3 & 43.9 & 67.8 & 63.5 & 65.4 & 79.4 & 43.2 & 27.2 & 57.1 \\ 
\hline
\text{Acos} & 17.29 & 77.4 & 43.6 & 70.1 & 69.8 & 64.4 & 77.2 & 41.4 & 52.5 & 62.0 \\
\text{Norm} & 17.76 & 77.3 & 43.5 & 70.4 & \textbf{71.5} & \textbf{66.4} & 77.4 & 40.6 & \textbf{54.6} & 62.7 \\
\text{JS} & 15.61 & \textbf{78.8} & \textbf{46.8} & \textbf{73.4} & 70.5 & 66.1 & \textbf{80.0} & \textbf{44.4} & 53.4 & \textbf{64.2} \\ 
\bottomrule
\end{tabular}
}
\label{tab:llama2_13b_25_ratio}
\end{table}

\begin{table}[h]
\centering
\caption{Performance on Llama2-13B at $40\%$ layer pruning ratio. Both SliceGPT and LLM-Pruner are applied with a desired $25\%$ sparsity ratio according to their implementation.}
\resizebox{\linewidth}{!}{
\begin{tabular}{c |c |c c c c c c c c c}
\toprule
Model & Wikitext & BoolQ & ARC-C & ARC-E & WG & HS & PIQA & OBQA & MMLU & Average \\ \hline
\text{Original} &12.96 & 79.8 & 49.0 & 77.2 & 71.6 & 69.6 & 81.3 & 45.2 & 55.0 & 66.1 \\ \hline
\text{ShortGPT} & 81.29 & 62.5 & 32.3 & 44.8 & 62.8 & 47.8 & 62.4 & 35.4 & 42.6 & 48.8 \\
\text{DeeperLayers} & 205.75 & 43.5 & 33.5 & 45.8 & 65.7 & 51.8 & 64.5 & 33.4 & 46.1 & 48.0 \\
\text{SliceGPT$^{*}$} & 37.06 & 37.5 & 29.2 & 44.1 & 61.6 & 49.6 & 59.9 & 35.6 & 27.3 & 43.1 \\ 
\text{LLM-Pruner$^{*}$} & \textbf{31.86} & 43.4 & 35.4 & \textbf{56.3} & 57.8 & \textbf{60.2} & \textbf{75.3} & \textbf{41.6} & 23.8 & 49.2 \\ 
\hline
\text{Acos} &38.65 & 57.5 & \textbf{36.0} & 53.0 & 65.4 & 54.9 & 69.8 & 33.6 & \textbf{49.6} & 52.5 \\
\text{Norm} & 38.65 & 57.5 & 36.0 & 53.0 & 65.4 & 54.9 & 69.8 & 33.6 & 49.6 & 52.5 \\
\text{JS} & 35.89 & \textbf{65.0} & 32.6 & 51.7 & \textbf{66.7} & 57.3 & 68.6 & 35.6 & 46.0 & \textbf{52.9} \\ 
\bottomrule
\end{tabular}
}
\label{tab:llama2_13b_40_ratio}
\end{table}

\begin{figure}[h]
\centering
\begin{subfigure}[b]{0.4\textwidth}
    \includegraphics[width=\columnwidth]{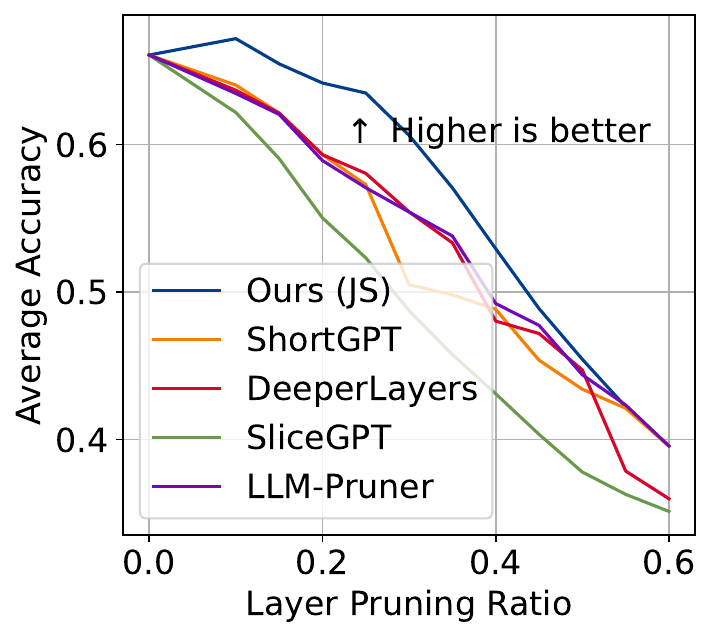}
    \caption{}
    \label{fig:acc_llama2_13b}
\end{subfigure}
\hspace{1cm}
\begin{subfigure}[b]{0.4\textwidth}
    \includegraphics[width=\columnwidth]{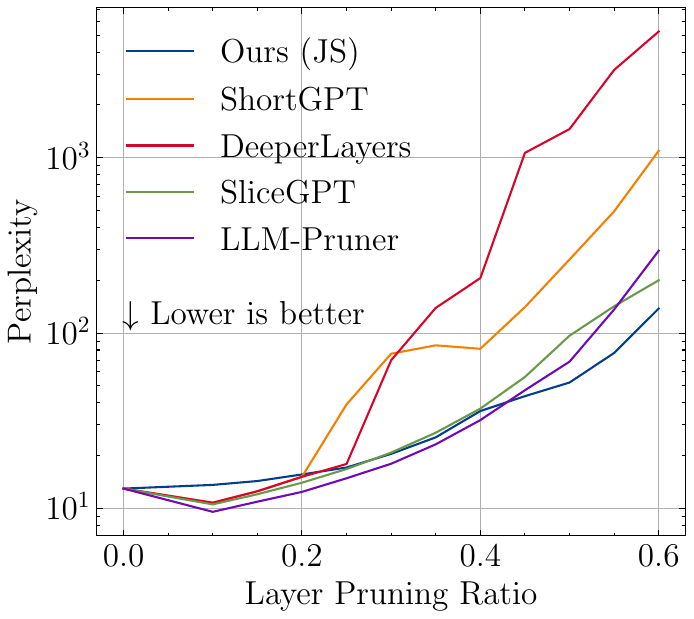}
    \caption{}
    \label{fig::ppl_llama2_13b}
\end{subfigure}
\caption{\textbf{(a)}: Average zero/few-shot performance at different layer pruning ratios on Llama2-13B. \textbf{(b)}: Perplexity (with a logarithmic scale) on WikiText2 at varying layer pruning ratios. \proj{} outperforms other methods including LLM-Pruner (not evaluated on Llama3 models) on both text generation and QA tasks. }
\label{fig:llama2_13b}
\end{figure}

\cref{fig:llama2_13b} shows the performance change at various layer pruning ratios. 

\section{Complementary results for Figure~\ref{fig:pruned_layers_overview}}
In this section, we provide two additional tables listing the pruned layers for readers interested in a more detailed breakdown of our pruning results. This table complements Fig. 4 by offering specific insights into which layers were pruned and the type of pruned layers (denoted with distinct colors). By including this supplementary information, we aim to enhance the transparency and reproducibility of our results, allowing for a deeper understanding of the pruning strategy employed in our study. Interested readers can use these results to reproduce pruned models for their purposes without running our code for pruning.

\begin{table}[t]
\centering
\begin{tabular}{c c c}
\toprule
Layer pruning ratio & Pruned Layers & Performance Drop \\ \hline
25\% & \attnbox{A13} \attnbox{A15} \attnbox{A18-19} \trbox{T20} \attnbox{A21-24} \trbox{T25} \attnbox{A26-27} \trbox{T28} & 11.6\% \\ 
\multirow{2}{*}{40\%} & \attnbox{A13} \mlpbox{F14} \attnbox{A15} \mlpbox{F16} \trbox{T17-20} & \multirow{2}{*}{33.2\%} \\ 
 & \attnbox{A21} \trbox{T22-23} \attnbox{A24} \trbox{T25-26} \attnbox{A27} \trbox{T28} & \\
\bottomrule
\end{tabular}
\caption{Pruned layers in Llama-3-8B at various layer pruning ratios. $A$ stands for a self-attention layer, $F$ denotes an FFN layer. And $T$ denotes a transformer block. }
\label{tab:pruned_layers_llama3_8b}
\end{table}

\begin{table}[t]
\centering
\begin{tabular}{c c c}
\toprule
Layer pruning ratio & Pruned Layers & Performance Drop \\ \hline
\multirow{2}{*}{25\%} & \attnbox{A33} \attnbox{A40-A50} \trbox{T51-52} \attnbox{A53-A57} \trbox{T58} & \multirow{2}{*}{1.9\%} \\ 
 & \trbox{T59} \attnbox{A60-A66} \trbox{T67} \attnbox{A68-A70} \mlpbox{F73} \attnbox{A74} \attnbox{A79}  &  \\ 
\multirow{2}{*}{40\%} & \attnbox{A33} \mlpbox{F35} \attnbox{A36} \attnbox{A38-40} \trbox{T41-44} \attnbox{A45} \trbox{T46-55} \attnbox{A56} & \multirow{2}{*}{17.5\%} \\ 
 & \trbox{T57-T63} \attnbox{A64} \trbox{T65-68} \attnbox{A69-70} \mlpbox{F73} \attnbox{A74} \attnbox{A79} & \\
\bottomrule
\end{tabular}
\caption{Pruned layers in Llama3-70B at various layer pruning ratios. $A$ stands for a self-attention layer, $F$ denotes an FFN layer. And $T$ denotes a transformer block.}
\label{tab:pruned_layers_llama3_70b}
\end{table}


\section{Text generation results on pruned models (w/o finetuning). }\label{app:text_generation}
Below, we provide generated samples for readers to have a glance at the generation ability after pruning.
\begin{table}[htbp]
\caption{Generated examples from the pruned LLama3-70B using \proj{}. The underlined texts denote the input prompts.}
\label{tab:gen_example_llama3_70b_part_1}
\begin{tabularx}{\textwidth}{l|X}
\toprule
Layer pruning ratio & Generated sentences \\ \hline
\multirow{3}{*}[-10em]{Original} & \underline{Neural Sequential Model, especially transformers,} have been widely used in NLP tasks. The sequential model is mainly based on the self-attention mechanism, which is used to capture the relationship between each token in the input sequence. The core of the self-attention mechanism is to calculate the similarity between the token in the input sequence, and then use the similarity to calculate the weight between each token. The self-attention mechanism can be divided into two types: local self-attention and global self-attention. Local self-attention only calculates the similarity between the token in the input sequence, while global self-attention calculates the similarity between the token in the input sequence and the token in the output sequence. The self-attention mechanism can be used to capture the relationship between each token in the input sequence, which is very important for NLP tasks. \\ 
\cmidrule{2-2}
 & \underline{Last night we all danced together in the rain,} the joy of the moment was so present that we didn’t want it to end. The raindrops were the perfect accent to the rhythm of the music. The night was perfect. We danced
in the rain. We danced for the rain. We danced for the joy of dancing in the rain. We danced for the joy of being alive, together, in the rain. \\ 
\cmidrule{2-2}
& \underline{It fills me with such pride and joy} to see the work that our team has done to build a company that is committed to serving the people of our community. I am so proud of what we have accomplished and the legacy we have built.
The first thing that I noticed about our company is that it is a family business. We are a family of four, and we are all very close.\\
\midrule
\multirow{3}{*}[-15em]{0.25} & \underline{Neural Sequential Model, especially transformers,} have been widely used for many applications in natural language processing. However, it is still difficult for most of the neural models to incorporate structured external knowledge, which is essential for some tasks. This paper presents a new approach to integrating structured knowledge into sequential models. We introduce a new type of recurrent neural network, which is inspired from the transformer model. We first introduce a new type of neural network called a recurrent neural network (RN). We show that this architecture is able to capture the structure of the data in an end-to-end manner. The proposed method is eval
uated on a large real-world data set of news articles. Experimental results show that our model outperforms the baseline in terms of accuracy and computation time.\\
\cmidrule{2-2}
& \underline{Last night we all danced together in the rain,} and I thought of you. The kids were playing in the street, and I thought of you. As I watched the sunset, I thought of you. When the sun rose this morning, I thought o
f you. I thought of you yesterday, I think of you today. I'll think of you again tomorrow. I have thought of you between the times I think I think of you. My thoughts are full of you. There is no room for another. You are my all, an
d I am yours. It's time to come home. I'm ready. I've been ready, and waiting. The light is green, my love. The coast is clear. \\
\cmidrule{2-2}
& \underline{It fills me with such pride and joy} to know that my son is being taught the way I want him to be taught. He is happy at school and is eager to go every day. He is learning about the Bible and how to be a good Chris
tian. I have complete confidence that he is in great hands at St. John’s. I am so thankful for the teachers and staff that help in the education and development of my child. Thank you for all you do!St. John’s has been a wonderful place for my children to grow and learn. They have been taught Christian values in a caring, nurturing environment. The teachers are excellent and the curriculum is great. \\
\bottomrule
\end{tabularx}
\end{table}

\begin{table}[htbp]
\caption{Generated examples from the pruned LLama3-70B using \proj{}. The underlined texts denote the input prompts.}
\label{tab:gen_example_llama3_70b_part_2}
\begin{tabularx}{\textwidth}{l|X}
\toprule
Layer pruning ratio & Generated sentences \\ \hline
\multirow{3}{*}[-15em]{0.4} & \underline{Neural Sequential Model, especially transformers,} are being applied in a diverse range of applications, from the traditional NLP tasks such as natural language generation, to the more general problems such as human
 relationship classification. In this workshop, we will review the recent advances and applications of transformers in a wide range of applications.
Neural Sequential Model, especially transformers, have been widely applied in a diverse range of applications, from the traditional NLP tasks such as natural language generation, to the more general problems, such as human relations
hip and interaction. In this workshop, we will review the recent advances and applications of transformers in a wide range of applications.
The first part of this workshop will be devoted to the basics of transformers, including the history and evolution of the transformers. The second part of this workshop will be devoted to the applications of transformers, including
the human relationship and human interaction. The last part of this workshop will be devoted to the future of transformers, including the future of the human interaction. The third part of this workshop will be devoted to the future
 of the human interaction. \\
\cmidrule{2-2}
 & \underline{Last night we all danced together in the rain,} the 3rd of 3 days of the 2018 Sante Festival. The final day of the festival was dedicated to the "Sante," a ceremonial blessing to the water, and a blessing for the pe
ople. This is a time where the people of the tribe are welcomed to come and be blessed by the monks. As the monk chants and blesses, a ripple of people come to be blessed, and then they in-in-in-in... in an alternating rhythm - just
 as the monks chants, and the music. It is a ritual that takes hours, and is a spiritual experience for all. \\
\cmidrule{2-2}
& \underline{It fills me with such pride and joy} when I come across a young man who has taken the initiative of his life. He is a man who is determined to be successful and to be a leader and not a victim. A man who has a desire to make the greatest use of the most important tool he possesses: his mind. I was watching TV when I was interviewed by a journalist. The story that attracted me was the story of a man who had been given the opportunity to enter the world of entrepreneurship and the results that he was experiencing.\\
\midrule
\multirow{3}{*}[-10em]{0.55} & \underline{Neural Sequential Model, especially transformers,} have been proven to be a powerful tool for the language learning. The combination of the recent advances in the field and the current pandemic circumstances, is an
excellent opportunity for me to build the first version of the 2023. by: 2023
 2023: the first year. I have been looking for the first time. \\
\cmidrule{2-2}
& \underline{Last night we all danced together in the rain,} while the sun was in the sky. And the rain was a joyful dance, the rain was the music. I think of the time of the sun and I of the wind, and the rain of the earth, is a gives all its power and strength.
I am in the midst of a multitude and the sun and the rain, we shall be at the same time. The sun and the water, is the power and the life and the air, and the rain is the music. \\
\cmidrule{2-2}
& \underline{It fills me with such pride and joy} that I am a product of the most powerful and effective Bible curriculum, the Catholic Church, of the Catholic Church. You are the product of the Catholic Church, and I am proud of the Catholic Church, and I am proud of the Catholic Church. The Catholic Church is a sacred and holy sacram, and it is a sacred sacram, which is a holy sacram. \\
\bottomrule
\end{tabularx}
\end{table}

\begin{table}[htbp]
\caption{Generated examples from the pruned LLama3-8B using \proj{}. The underlined texts denote the input prompts.}
\label{tab:gen_example_llama3_8b_part_1}
\begin{tabularx}{\textwidth}{l|X}
\toprule
Layer pruning ratio & Generated sentences \\ \hline
\multirow{3}{*}[-15em]{Original} & \underline{Neural Sequential Model, especially transformers,} are the state-of-the-art models for various NLP tasks. However, these models are often too large for practical deployment, due to the huge number of parameters and memory consumption. For example, BERT-Base has 110M parameters and takes 500M memory. The large model size and memory consumption limit their applications to mobile devices, which are often memory constrained and resource limited. In this paper, we propose a novel and efficient method, named model compression with knowledge distillation (MCKD), to compress the large models with knowledge distillation. Specifically, we first train a small teacher model, and the
n train a student model with the knowledge from the teacher model. We find that a student model with 1/100 parameters and 1/1000 memory can achieve comparable performance with the teacher model. We also show that the student model c
an be further compressed by removing unnecessary parameters. We evaluate our MCKD method on various NLP tasks, including text classification, sequence labeling, and machine translation, and demonstrate that the proposed MCKD method
can significantly improve the performance of the student model. \\
\cmidrule{2-2}
& \underline{Last night we all danced together in the rain,} under a full moon, on the beach. It was one of those nights that you always hope for, but never really think will happen. The stars were out, the moon was huge and ful
l, the ocean was calm, and it was warm enough to stay outside and not freeze to death. We made s’mores, we danced, we laughed, we talked, we listened to music, and we had a great time. It was the kind of night that you can’t help bu
t smile at as you lay in bed, drifting off to sleep. The kind of night that makes you feel like anything is possible. The kind of night that makes you feel alive and like you are living in the moment. \\
\cmidrule{2-2}
& It fills me with such pride and joy to be able to share these photos with you. I am a mother to four boys. I am an artist and designer. I am a wife. I am a friend. I am a daughter. I am a sister. I am a writer. I am a baker. I am a cook. \\
\midrule
\multirow{3}{*}[-10em]{0.25} & \underline{Neural Sequential Model, especially transformers,} is a powerful tool that has been very effective in generating text and music, and other artificial intelligence tasks. In the original Transformer, the Transformer
block was a fully connected layer of the neural network, which is a dense layer and is the default setting in the neural sequential model. It is a very powerful model, but the problem is that the original Transformer is too time-consuming, which is a problem that the original Transformer is a bit of a problem.
So, the Transformer model is now available as a module on the Open Neural Networks Library (C code) and a C++ version of the original Transformer. The TensorFlow implementation of the Neural Sequential Model is a good choice for tho
se who are interested in the Neural Sequential Model. \\
\cmidrule{2-2}
& \underline{Last night we all danced together in the rain,} and the rain poured on the dance floor as the crowd of 1,200, 3D Audio, and the 1971 film The Paper Man, a documentary, were all together with the music of the 1971 film The Godfather. The music is also a classic, and the music is a classic, the same as the 1971 film The Godfather, a documentary. \\
\cmidrule{2-2}
& \underline{It fills me with such pride and joy} to hear the words “thank you for the book, thank you for the book” for the first time. A copy of my book is now in the hands of a little girl, a little girl who was born in 2012, and a little girl who is not the only one who is not the only one to the book. The book is the 2012 release of the book. The book is the first book in the "Pandora” trilogy trilogy, which is the second in a trilogy \\
\bottomrule
\end{tabularx}
\end{table}

\begin{table}[htbp]
\caption{Generated examples from the pruned LLama3-8B using \proj{}. The underlined texts denote the input prompts.}
\label{tab:gen_example_llama3_8b_part_2}
\begin{tabularx}{\textwidth}{l|X}
\toprule
Layer pruning ratio & Generated sentences \\ 
\hline
\multirow{3}{*}[-5em]{0.40} & \underline{Neural Sequential Model, especially transformers,} has the the ability to capture the data of the new, and the Leberton, for a new, and the the ability to capture the data of the the ability to the the the the the and the the the and the the the the and the the. \\
\cmidrule{2-2}
& \underline{Last night we all danced together in the rain,} and in the clouds of the rain, it was the rain and the rain and the rain. This is the story of the rain, and the rain and the rain.
The rain was a wonderful gift. \\
\cmidrule{2-2}
& \underline{It fills me with such pride and joy} that I am now considered as a new and it's a new brand. The name is the name of a new Brand. I have a good name for the name. I will be a new brand is a new brand. I have a good
name for the name. I will be a new
I want to be a new brand. I want to be a new brand. I want to be a new brand. I want to be a new brand. \\
\midrule
\multirow{3}{*}[-3em]{0.55} & \underline{Neural Sequential Model,} especially transformers, is a source of the German- hering, the German-7. The 2nd. The 10th of 20 the 22. 2. The 4. 2. 3 12. 2. 3 12. \\
\cmidrule{2-2}
& \underline{Last night we all danced together in the rain,} that was the first of the two. The first of the two.
The next one is one of the three, in the is one of the three, and the another one is one of the the four. \\
\cmidrule{2-2}
& \underline{It fills me with such pride and joy,} I don't do the country, Toby Franey. And I don't dun credit I have the 4 of the 12 pack or whatever have you to I have the the. \\
\bottomrule
\end{tabularx}
\end{table}


\end{document}